\pdfoutput=1

\documentclass[11pt]{article}
\usepackage{authblk}

\usepackage{EMNLP2022}
\usepackage{booktabs}

\usepackage{times}
\usepackage{latexsym}
\usepackage{linguex}
\usepackage{enumitem}
\usepackage{algorithm}
\usepackage{algpseudocode}
\usepackage{graphicx}
\usepackage[british]{babel}

\usepackage{amsmath}

\usepackage[T1]{fontenc}

\usepackage[utf8]{inputenc}

\usepackage{microtype}

\usepackage{inconsolata}
\usepackage{multirow}
\usepackage{hyperref}

\DeclareMathOperator{\train}{train}
\DeclareMathOperator{\test}{test}
\DeclareMathOperator{\length}{length}
\DeclareMathOperator{\distance}{distance}

\title{The Better Your Syntax, the Better Your
Semantics? Probing Pretrained Language Models for the English Comparative Correlative}

\author[*$\diamond$]{Leonie Weissweiler}
\author[$\dag$*]{Valentin Hofmann}
\author[*$\diamond$]{Abdullatif Köksal}
\author[*$\diamond$]{Hinrich Sch\"utze}

\affil[*]{Center for Information and Language Processing, LMU Munich}
\affil[$\diamond$]{Munich Center of Machine Learning}
\affil[$\dag$]{Faculty of Linguistics, University of
Oxford \protect\\
\texttt{\{weissweiler,akoksal\}@cis.lmu.de} \protect\\ \texttt{valentin.hofmann@ling-phil.ox.ac.uk}}

\begin{document}
\maketitle
\begin{abstract}
Construction Grammar (CxG) is a paradigm from cognitive linguistics emphasising the connection between syntax and semantics. Rather than rules that operate on lexical items, it posits \textit{constructions} as the central building blocks of language, i.e., linguistic units of different granularity that combine syntax and semantics. As a first step towards assessing the compatibility of CxG with the syntactic and semantic knowledge demonstrated by state-of-the-art pretrained language models (PLMs), we present an investigation of their capability to classify and understand one of the most commonly studied constructions, the English comparative correlative (CC). We conduct experiments examining the classification accuracy of a syntactic probe on the one hand and the models' behaviour in a semantic application task on the other, with BERT, RoBERTa, and DeBERTa as the example PLMs. 
Our results show that all three investigated PLMs are able
to recognise the structure of the CC but fail to use its
meaning.
While human-like performance of PLMs on many NLP tasks has been alleged, this indicates that PLMs still suffer from substantial shortcomings in central domains of linguistic knowledge.

\end{abstract}

\setlength{\Exlabelsep}{0pt}
\setlength{\Extopsep}{5pt}

\section{Introduction}

The sentence ``The better your syntax, the better your
semantics.''  contains a construction called the English
comparative correlative
(CC; \citealp{fillmore1986}). Paraphrased, it could be read
as ``If your syntax is better, your semantics will also be
better.'' Humans reading this sentence are capable of doing
two things: (i) \textit{recognising} that two instances of
``the'' followed by an adjective/adverb in the comparative
as well as a phrase of the given structure (i.e., the syntax
of the CC) express a specific meaning (i.e., the semantics
of the CC); (ii) \textit{understanding}
the semantic meaning conveyed by the CC, i.e., understanding
that in a sentence
of the given structure, the second half is somehow
correlated with the first.

In this paper, we ask the following question: are pretrained
language models (PLMs) able to achieve these two steps?
This question is important for two reasons. Firstly, we hope
that recognising the CC and understanding its meaning is
challenging for PLMs, helping to set the research agenda for
further
improvements. Secondly, the CC is one of the most commonly
studied constructions in construction grammar (CxG), a
usage-based syntax paradigm from cognitive linguistics, thus
providing an interesting alternative to the currently
prevailing practice of analysing the syntactic capabilities
of PLMs with theories from generative grammar
(e.g., \citealp{marvin-linzen-2018-targeted}).

We divide our investigation into two parts. In the first
part, we examine the CC's syntactic properties and how
they are represented by PLMs, with the objective to
determine whether PLMs can \textit{recognise} an instance of
the CC. More specifically, we construct two syntactic probes
with different properties: one is inspired by recent probing
methodology (e.g., \citealp{belinkov-etal-2017-neural,
conneau-etal-2018-cram}) and draws upon minimal pairs to
quantify the amount of information contained in each PLM
layer; for the other one, we write a context-free grammar
(CFG) to construct approximate minimal pairs in which only
the word order determines if the sentences are an instance
of the CC or not. We find that starting from the third layer, all
investigated PLMs are able to distinguish positive from
negative instances of the CC. However, this method only
covers one specific subtype of comparative sentences. To cover the full diversity of instances, we conduct
an additional experiment for which we collect and manually label sentences from C4 \cite{raffel2020} that
resemble instances of the CC, resulting in a diverse set of
sentences that either are instances of the CC or resemble
them closely \textit{without} being instances of the CC. Applying the same methodology to this set of sentences, we observe that all examined PLMs are still able to separate the examples very well. 

In the second part of the paper, we aim to determine if the PLMs are able to \textit{understand} the meaning of the CC. We generate test scenarios in which a statement containing the CC is given to the PLMs, which they then have to apply in a zero-shot manner.
As this way of testing PLMs is prone to a variety of biases,
we introduce several mitigating methods in order to
determine the full capability of the PLMs. We find that none of the PLMs we investigate perform
above chance level, indicating that they are not able to
understand and apply the CC in a measurable way in this context.

We make three main \textbf{contributions}:
\begin{itemize}[leftmargin=*]
\item[--] We present the first comprehensive study examining how well PLMs can recognise and
understand a CxG construction, specifically the English comparative correlative.
\item[--] We develop a way of testing the PLMs' recognition of the CC that overcomes the challenge of probing for linguistic phenomena not lending themselves to minimal pairs.
\item[--] We adapt methods from zero-shot prompting and calibration to develop a way of testing PLMs for their understanding of the CC.\footnote{In order to foster research at the intersection of NLP and construction grammar, we will make our data and code available at \url{https://github.com/LeonieWeissweiler/ComparativeCorrelative}.}
\end{itemize}

\section{Construction Grammar}

\subsection{Overview}

A core assumption of generative grammar \citep{chomsky1988}, which can be already found in Bloomfieldian structural linguistics 
\citep{Bloomfield.1933}, is a strict separation 
of lexicon and grammar: grammar is conceptualized as a set
of compositional and general rules that operate on a list
of arbitrary and specific lexical items in generating
syntactically well-formed sentences. This dichotomous view
was increasingly questioned in the 1980s when several
studies drew attention to the fact that linguistic units
larger than lexical items (e.g., idioms) can also possess
non-compositional meanings \citep{Langacker.1987,
Lakoff.1987, Fillmore.1988, Fillmore.1989}.
For instance, it is not clear how the effect of the words ``let alone''(as in ``she doesn't eat fish,
let alone meat'') on both the syntax and the semantics of the rest of the sentence could be inferred from general syntactic rules \citep{Fillmore.1988}..
This insight about the ubiquity of stored form-meaning
pairings in language is
adopted as the central tenet
of
grammatical theory by Construction Grammar (CxG;
see \citet{hoffmann2013oxford} for a comprehensive
overview). Rather than a system divided into non-overlapping
syntactic rules and lexical items, CxG views language as a
structured system of constructions with varying
granularities that encapsulate syntactic and semantic
components as single linguistic signs---ranging from
individual morphemes up to phrasal elements and fixed
expressions \citep{Kay.1999, Goldberg.1995}. In this
framework, syntactic rules can be seen as emergent
abstractions over similar stored
constructions \citep{goldberg2003, Goldberg.2006}. A
different set of stored constructions can result in
different abstractions and thus different syntactic rules,
which allows CxG to naturally accommodate for the dynamic
nature of grammar as evidenced, for instance, by
inter-speaker variability and linguistic
change \citep{Hilpert.2006}.

\subsection{Construction Grammar and NLP}
We see three main motivations for the development of a first probing approach for CxG:
\begin{itemize}[leftmargin=*]
    \item[--] We believe that the active discourse in (cognitive) linguistics about the best description of human language capability can be supported and enriched through a computational exploration of a wide array of phenomena and viewpoints. We think 
    that the probing literature in NLP investigating linguistic phenomena with computational methods should be diversified to include theories and problems from all points on the broad spectrum of linguistic scholarship. 
    \item[--] We hope that the investigation of large PLMs'
    apparent capabilities to imitate human language and the
    mechanisms responsible for these capabilities will be enriched by
    introducing a usage-based approach to grammar. This is
    especially important as some of the discourse in recent
    years has focused on the question of whether PLMs are
    constructing syntactically acceptable sentences for the
    correct reasons and with the correct underlying
    representations (e.g. \citealp{mccoy-etal-2019-right}). We would like to suggest that
    considering alternative theories of grammar,
    specifically CxG with its
    incorporation of slots in constructions that may be
    filled by specific word types and its focus on
    learning
    without an innate, universal grammar, may be beneficial to understanding the learning process of PLMs as their capabilities advance further.
    \item[--] Many constructions present an interesting challenge for PLMs. In fact, recent work in challenge datasets \cite{ribeiro-etal-2020-beyond} has already started using what could be considered constructions, in an attempt to identify types of sentences that models struggle with, and to point out a potential direction for improvement. One of the central tenets of CxG is the relation between the form of a construction and its meaning, or to put it in NLP terms, a model must 
    learn to infer parts of the sentence meaning from patterns that are present in it, as opposed to words.
    We believe this to be an interesting challenge for future PLMs.
\end{itemize}

\subsection{The English Comparative Correlative}
The English comparative correlative (CC) is one of the most commonly studied constructions in linguistics, for several reasons. Firstly, 
it constitutes a clear example of a linguistic phenomenon
that is challenging to explain
in the framework of generative grammar
\citep{culicover1999, abeille2008}, even though
there have been approaches following that school of
thought \citep{dikken2005, iwasaki2009}. Secondly, it exhibits a range of interesting syntactic and semantic features, as detailed below. 
These reasons, we believe, also make the CC an ideal testbed for a first study attempting to extend the current trend of syntax probing for rules by developing methods for probing according to CxG.

The CC can take many different forms, some of which are exemplified here:

\ex.\label{cc_1} The more, the merrier.

\ex.\label{cc_2} The longer the bake, the browner the colour.

\ex.\label{cc_3} The more she practiced, the better she became.

Semantically, the CC consists of two clauses, where the second clause can be seen as the dependent variable for the independent variable specified in the first one \cite{goldberg2003}. It can be seen on the one hand as a statement of a general cause-and-effect relationship, as in a general conditional statement (e.g., \ref{cc_2} could be paraphrased as ``If the bake is longer, the colour will be more brown''), and on the other as a temporal development in a comparative sentence (paraphrasing \ref{cc_3} as ``She became better over time, and she practiced more over time''). Usage of the CC typically implies both readings at the same time.
Syntactically, the CC is characterised in both clauses by an
instance of ``the'' followed by an adverb or an adjective in
the comparative, either with ``-er'' for some adjectives
and adverbs, or with ``more'' for others, or special
forms like ``better''.
Special features of the comparative sentences following this
are the optional omission of the future ``will'' and of
``be'', as in \ref{cc_1}. Crucially, ``the'' in this
construction does not function as a determiner of noun
phrases \cite{goldberg2003}; rather, it has a function specific to the CC and has variously been called a ``degree word'' \cite{dikken2005} or ``fixed material'' \cite{hoffmann2019more}.

\section{Syntax}

Our investigation of PLMs' knowledge of the CC is split into
two parts.  First, we probe for the PLMs' knowledge of the
syntactic aspects of the CC, to determine if they recognise
its structure. Then we devise a test of their understanding
of its semantic aspects by investigating their ability to
apply, in a given context, information conveyed by a CC.

\subsection{Probing Methods}

As the first half of our analysis of PLMs' knowledge of the CC, we investigate its syntactic aspects. Translated into probing questions, this means that we ask: can a PLM recognise an instance of the CC? Can it distinguish instances of the CC from similar-looking non-instances? Is it able to go beyond the simple recognition of its fixed parts (``The \texttt{COMP-ADJ}/\texttt{ADV}, the ...'') and group all ways of completing the sentences that are instances of the CC separately from all those that are not? And to frame all of these questions in a syntactic probing framework: will we be able to recover, using a logistic regression as the probe, this distinguishing information from a PLM's embeddings?

The established way of testing a PLM for its syntactic knowledge has in recent years become minimal pairs (e.g., \citealp{warstadt-etal-2020-blimp-benchmark}, \citealp{Demszky.2021}). This would mean pairs of sentences which are indistinguishable except for the fact that one of them is an instance of the CC and the other is not, allowing us to perfectly separate a model's knowledge of the CC from other confounding factors. While this is indeed possible for simpler syntactic phenomena such as verb-noun number agreement,
there is no obvious way to construct minimal pairs for the CC.
We therefore construct minimal pairs in two ways: one with artificial data based on a context-free grammar (CFG), and one with sentences extracted from C4.

\subsubsection{Synthetic Data}

In order to find a pair of sentences that is as close as
 possible to a minimal pair, we devise a way to modify the
 words following ``The X-er'' such that the sentence is no
 longer an instance of the construction. The pattern for a
 positive instance is ``The \texttt{ADV}-er
 the \texttt{NUM} \texttt{NOUN} \texttt{VERB}'', e.g., ``The
 harder the two cats fight''.
To create a negative instance,
we reorder the pattern to
 ``The \texttt{ADJ}-er \texttt{NUM} \texttt{VERB}
 the \texttt{NOUN}'', e.g., ``The harder two fight the
 cats''. The change in role of the numeral from the
 dependent of a head to a head itself, made possible by
 choosing a verb that can be either transitive or
 intransitive, as well as the change from an adverb to an
 adjective, allows us to construct a negative instance that
 uses the same words as the positive one, but in a different
 order.\footnote{Note that an alternative reading of this
 sentence exists: the numeral ``two'' forms the noun phrase by
 itself and ``The harder'' is still interpreted as part of
 the CC. The sentence is actually a positive instance on
 this interpretation.
We regard this reading as very improbable.} In order to
 generate
 a large number of instances, we collect two sets each of
 adverbs, numerals, nouns and verbs that are mutually
 exclusive between training and  test sets. To investigate if the model is confused by additional content in the sentences, we write an CFG to insert phrases before the start of the first half, in between the two halves, and after the second half of the CC (see Appendix, Algorithms~\ref{cfg1} and \ref{cfg2} for the complete CFG).

While this setup is rigourous in the sense that positive and negative sentences
are exactly matched, it comes with the drawback of only considering one type of CC. 
To be able to conduct a more comprehensive investigation, we 
adopt a complementary approach and
turn to
pairs extracted from C4 (see Appendix,
Tables \ref{tab:examples_art_train}
and \ref{tab:examples_art_test}, for examples of training
and test data).
These cover a broad range of CC patterns, albeit
without meeting the criterion that positive and negative
samples are exactly matched.

\subsubsection{Corpus-based Minimal Pairs}

While accepting that positive and negative instances
extracted from a corpus will automatically not be minimal and
therefore contain some lexical overlap and context cues, we
attempt to regularise our retrieved instances as far as
possible. To form a first candidate set, we POS tag C4 using
spaCy \cite{spacy_2018} and extract all sentences that
follow the pattern ``The'' (\texttt{DET}) followed by either
``more'' and an adjective or adverb, or an adjective or
adverb ending in ``-er'', and at any point later in the
sentence again the same pattern. We discard examples with
adverbs or adjectives that were falsely labelled as
comparative, such as ``other''. We then group these
sentences by their sequence of POS tags, and manually classify the sequences as either positive or negative instances.
We observe that sentences sharing a POS tag pattern tend to be either all negative or all positive instances, allowing us to save annotation time by working at the POS tag pattern level instead of the sentence level. To make the final set as diverse as possible, we sort the patterns randomly and label as many as possible. In order to further reduce  interfering factors in our probe, we separate the POS tag patterns between training and  test sets (see Appendix, Table \ref{tab:examples_corpus}, for examples).

\subsubsection{The Probe}

For both datasets, we investigate the overall accuracy of our probe as well as the impact of several factors. The probe consists of training a simple logistic regression model on top of the mean-pooled sentence embeddings \citep{Vulic.2020}. 
To quantify the impact of the length of the sentence, the
start position of the construction, the position of its
second half, and the distance between them, we construct
four different subsets $D^{\train}_{f}$ and $D^{\test}_{f}$
from both the artificially constructed and the corpus-based dataset.
For each subset, we sample sentences such that both the positive and the negative class is balanced across every value of the feature within a certain range of values. 
This ensures that the probes are unable to exploit correlations between a class and any of the above features. We create the dataset as follows
	\[
	D_f=\bigcup_{v\in f_v} \bigcup_{l^{*}\in L} S(D,v,l^{*},n^{*}),
	\]
where $f$ is the feature, $f_v$ is the set of values for
	$f$, $L=\{positive, negative\}$ are the labels, and $S$ is a function
that returns $n^{*}$ elements from $D$ that have value $v$
	and label $l^{*}$.

To make this task more cognitively realistic, we aim to test if a model is able to generalise from shorter sentences, which contain relatively little additional information besides the parts relevant to the classification task, to those with greater potential interference due to more additional content that is not useful for classification.
Thus, we restrict the training set to samples from the lowest quartile of each feature so that 
 $f_v$ becomes $[v_f^{\min}, v_f^{\min}+\frac{1}{4}(v_f^{\max}-v_f^{\min})]$ for $D_f^{\train}$ and $[v_f^{\min}, v_f^{\max}]$ for $D_f^{\test}$.
We report the test performance for every value of a given feature separately to recognise patterns.
For the artificial syntax probing, we generate 1000 data points for each value of each feature for each training and test for each subset associated with a feature. For the corpus syntax probing, we collect 9710 positive and 533 negative sentences in total, from which we choose 10 training and 5 test sentences for each value of each feature in a similar manner.
To improve comparability and make the experiment computationally feasible, we test the ``large'' size of each of our three models, using the Huggingface Transformers library \cite{wolf2019}.  Our logistic regression probes are implemented using Scikitlearn \cite{scikit-learn}.

\subsection{Probing Results}

\begin{figure}
    \includegraphics[width=\columnwidth]{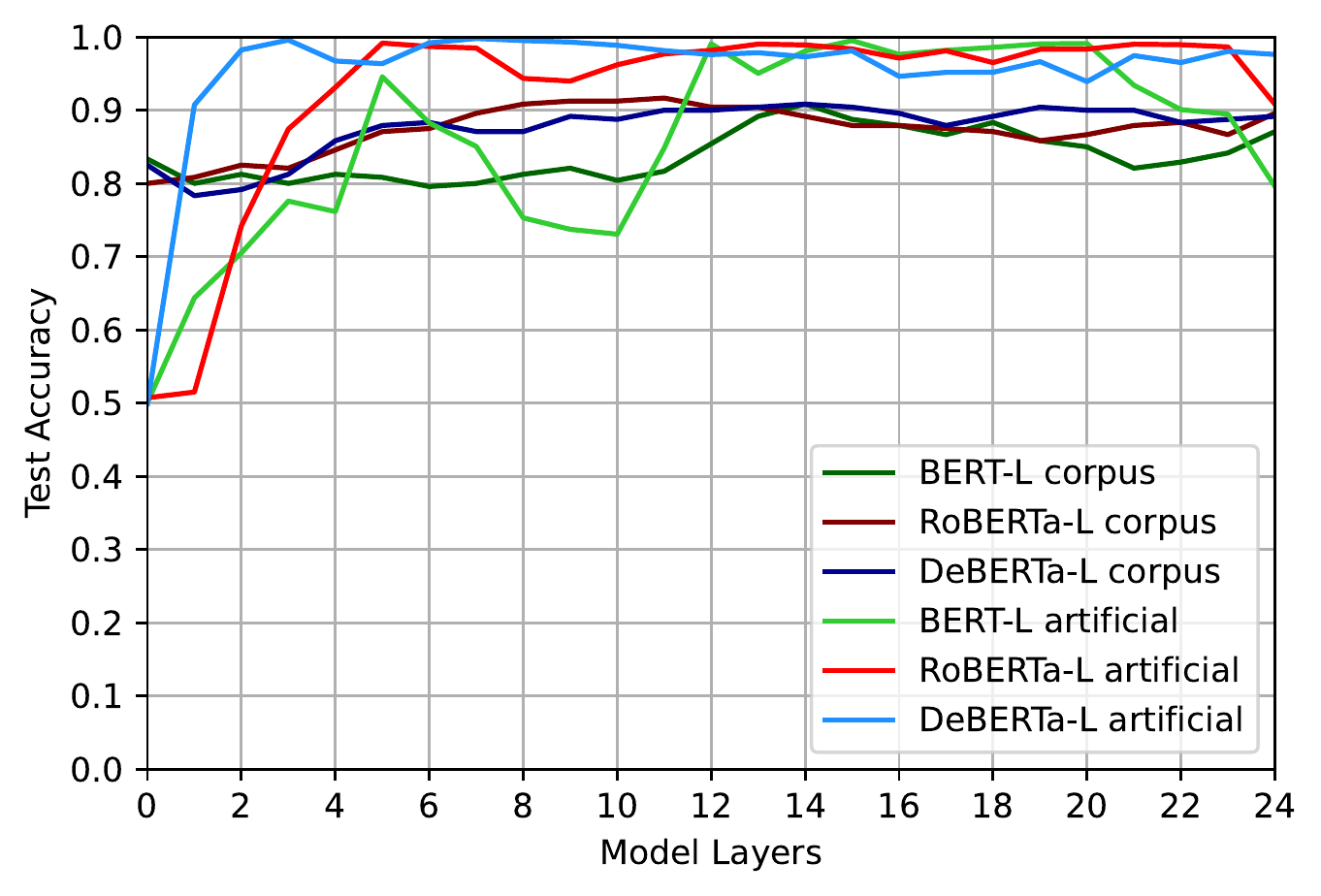}
    \caption{Overall accuracy per layer for $D_{\length}$.
    All shown models are the large model variants.
The models can easily distinguish
    between positive and negative examples in at least some
    of their layers.
}
    \label{syntax_results}
\end{figure}

\subsubsection{Artificial Data}

As shown in Figure \ref{syntax_results}, the results of our syntactic probe indicate that all models can easily distinguish between positive and negative examples in at least some of their layers, independently of any of the sentence properties that we have investigated. We report full results in the Appendix in Figures \ref{bert_artificial}, \ref{roberta_artificial}, and \ref{deberta_artificial}.
We find a clear trend that DeBERTa performs better than
RoBERTa, which in turn performs better than BERT across the
board.
As DeBERTa's performance in all layers is nearly
perfect, we are unable to observe patterns related to the length of the sentence, the start position of the CC, the start position of the second half
of the CC, and the distance between them. By contrast, we
observe interesting patterns for BERT and RoBERTa. For $D_{\length}$, and to a
lesser degree $D_{\distance}$ (which correlates with it), we
observe that at first, performance goes down with
increased length as we would expect---the model
struggles to generalise to longer sentences with more
interference since it was only trained on short ones. However, this trend is reversed in the last few layers. We hypothesize this may be due to an increased focus on semantics in the last layers \citep{Peters.2018, Tenney.2019}, which could lead to interfering features particularly in shorter sentences. 

\subsubsection{Corpus Data}

In contrast, the results of our probe on more natural data from C4 indicate two different trends: first, as the positive and negative instances are not identical on a bag-of-word level, performance is not uniformly at 50\% (i.e., chance) level in the first layers, indicating that the model can exploit lexical cues to some degree. We observe a similar trend as with the artificial experiment, which showed that DeBERTa performs best and BERT worst. The corresponding graphs can be found in the Appendix in Figures \ref{bert_corpus}, \ref{roberta_corpus}, and \ref{deberta_corpus}.

Generally, this additional corpus-based experiment validates our findings from the experiment with artificially generated data, as all models perform at 80\% or better from the middle layers on, indicating that the models are able to classify instances of the construction even when they are very diverse and use unseen POS tag patterns.

Comparing the average accuracies on $D_{\length}$ for both data sources in Figure \ref{syntax_results}, we observe that all models perform better on artificial than on corpus data from the fifth layer on, with the notable exception of a dip in performance for BERT large around layer 10. 

\section{Semantics}

\begin{table*}[]
\centering
\small
\begin{tabular}{@{}ccll@{}}
\toprule
No. & Purpose & Approach               & Sentence Schema                                                                                                                                             \\ \midrule
\multirow{2}{*}{S1}      & \multicolumn{2}{c}{\multirow{2}{*}{\hspace{-1.2em}Base}}                  & The \texttt{ADJ1}-er you are, the \texttt{ADJ2}-er you are. The \texttt{ANT1}-er you are, the \texttt{ANT2}-er you are.\\[0.2em]
& & & \texttt{NAME1} is \texttt{ADJ1}-er than \texttt{NAME2}. Therefore, \texttt{NAME1} is \texttt{[MASK]} than \texttt{NAME2}. \\[0.1em]
\midrule
\multirow{2}{*}{S2} & \multirow{7}{*}{\rotatebox{90}{\small {\begin{tabular}{c}Bias Test\end{tabular}}}} & \multirow{2}{*}{Recency}     & The \texttt{ANT1}-er you are, the \texttt{ANT2}-er you are. The \texttt{ADJ1}-er you are, the \texttt{ADJ2}-er you are.\\[0.2em]
& & & \texttt{NAME1} is \texttt{ADJ1}-er than \texttt{NAME2}. Therefore, \texttt{NAME1} is \texttt{[MASK]} than \texttt{NAME2}. \\[0.4em]
\multirow{2}{*}{S3} &   & \multirow{2}{*}{Vocabulary}  & The \texttt{ADJ1}-er you are, the \texttt{ANT2}-er you are. The \texttt{ANT1}-er you are, the \texttt{ADJ2}-er you are.\\
& & & \texttt{NAME2} is \texttt{ADJ1}-er than \texttt{NAME2}. Therefore, \texttt{NAME1} is \texttt{[MASK]} than \texttt{NAME2}. \\[0.4em]
\multirow{2}{*}{S4}   &   & \multirow{2}{*}{Name}    & The \texttt{ADJ1}-er you are, the \texttt{ADJ2}-er you are. The \texttt{ANT1}-er you are, the \texttt{ANT2}-er you are. \\[0.2em]
& & & \texttt{NAME2} is \texttt{ADJ1}-er than \texttt{NAME1}. Therefore, \texttt{NAME2} is \texttt{[MASK]} than \texttt{NAME1}.\\[0.1em]
\midrule
S5 &  \multirow{6}{*}{\rotatebox{90}{\small {\begin{tabular}{c}Calibration\end{tabular}}}} & Short     & \texttt{NAME1} is \texttt{ADJ1}-er than \texttt{NAME2}. Therefore, \texttt{NAME1} is \texttt{[MASK]} than \texttt{NAME2}.\\[0.4em]
\multirow{2}{*}{S6}  & & \multirow{2}{*}{Name}      & The \texttt{ADJ1}-er you are, the \texttt{ADJ2}-er you are. The \texttt{ANT1}-er you are, the \texttt{ANT2}-er you are. \\[0.2em]
&  & & \texttt{NAME1} is \texttt{ADJ1}-er than \texttt{NAME2}. Therefore, \texttt{NAME3} is \texttt{[MASK]} than \texttt{NAME4}.\\[0.4em]
\multirow{2}{*}{S7}   &   & \multirow{2}{*}{Adjective} & The \texttt{ADJ1}-er you are, the \texttt{ADJ2}-er you are. The \texttt{ANT1}-er you are, the \texttt{ANT2}-er you are. \\[0.2em]
& & & \texttt{NAME1} is \texttt{ADJ3}-er than \texttt{NAME2}. Therefore, \texttt{NAME1} is \texttt{[MASK]} than \texttt{NAME2}. \\[0.2em] \bottomrule
\end{tabular}
\caption{Overview of the schemata of all test scenarios used for semantic probing}
\label{semantics_schemata}
\end{table*}

\subsection{Probing Methods}

\subsubsection{Usage-based Testing}

For the second half of our investigation, we turn to
semantics. In order to determine if a model has understood
the meaning of the CC, i.e., if it has understood
that in any sentence, ``the \texttt{COMP}
.... the \texttt{COMP}'' implies a correlation between the
two halves, we adopt a usage-based approach and ask: can the
model,
based on the meaning conveyed by the CC,
draw a correct inference
in a specific scenario?
For
this, we
construct general test instances of the CC that consist of a
desired update of the belief state of the model about the
world, which we then expect it to be able to apply. More
concretely, we generate sentences of the form
``The \texttt{ADJ1}-er you are, the \texttt{ADJ2}-er you
are.'', while picking adjectives at random. To this general
statement, we then add a specific scenario with two random
names: ``\texttt{NAME1} is \texttt{ADJ1}-er
than \texttt{NAME2}.'' and ask the model to
draw an inference from it by predicting a token at the masked
position in the following sentence:
``Therefore, \texttt{NAME1} is \texttt{[MASK]}
than \texttt{NAME2}.'' If the model has understood the
meaning conveyed by the CC and is able to
use it in predicting the mask, we expect the
probability of \texttt{ADJ2} to be high. To provide the
model with an alternative, we add a second sentence, another instance
of the CC, using the antonyms of the two adjectives. This
sentence is carefully chosen to have no
impact on the best filler for \texttt{[MASK]}, but also for other
reasons explained in Section \ref{biases}.
The full test context is shown in Table \ref{semantics_schemata}, S1. This enables us to compare the probability of \texttt{ADJ2} for the mask token directly with a plausible alternative, \texttt{ANT2}. One of our test sentences might be ``The stronger you are, the faster you are. The weaker you are, the slower you are. Terry is stronger than John. Therefore, Terry will be \texttt{[MASK]} than John'', where we compare the probabilities of ``faster'' and ``slower''.

Note that success in our experiment does not necessarily
indicate that the model has fully understood the meaning of
the CC.
The experiment can only provide a lower bound 
for the underlying understanding of any model.
However, we believe that our task is not unreasonable for a masked language model in a zero-shot 
setting. It is comparable in difficulty and non-reliance on world knowledge to the NLU tasks presented in 
LAMBADA \citep{paperno-etal-2016-lambada}, on which GPT-2 (117M to 1.5B parameters) has achieved high zero-shot accuracy (\citet{radford2019}, Table 3). 
While we investigate masked language models and not GPT-2, our largest models are comparable in size to
the sizes of GPT-2
that were used
(340M for BERT\textsubscript{L}, 355M for RoBERTa\textsubscript{L}, and 1.5B parameters for DeBERTa-XXL\textsubscript{L}), and we
believe that this part of our task is achievable to some degree.

\subsubsection{Biases}
\label{biases}
In this setup, we hypothesise several biases that models could exhibit and might cloud our assessment of its understanding of the CC, and devise a way to test their impact.

Firstly, we expect that models might prefer to repeat the adjective that is closest to the mask token. This has recently been documented for prompt-based experiments \cite{pmlr-v139-zhao21c}. Here, this adjective is \texttt{ANT2}, the wrong answer. To test the influence this has on the prediction probabilities, we construct an alternative version of our test context in which we flip the first two sentences so that the correct answer is now more recent. The result can be found in Table \ref{semantics_schemata}, S2.

Secondly, we expect that models might assign higher
probabilities to some adjectives, purely based on their
frequency in the pretraining corpus, as for example observed
by \citet{holtzman-etal-2021-surface}. To test this, we
construct a version of the test context in which
\texttt{ADJ2}/\texttt{ANT2} are swapped,
which means that we can keep both the overall words the same
as well as the position of the correct answer, while
changing which adjective it is. The sentence is now S3 in
Table \ref{semantics_schemata}. If there is a large
difference between the prediction probabilities for the
two different
versions,
that
this
means that a model's prediction is
influenced by the lexical identity of the
adjective in question.

Lastly, a model might have learned to associate adjectives
with names in pretraining, so we construct a third version,
in which we swap the names. This is S4 in
Table \ref{semantics_schemata}. If any prior association
between names and adjectives influences the prediction, we
expect the scores between S4 and S1 to differ.

\subsubsection{Calibration}

After quantifying the biases that may prevent us from seeing
a model's true capability in understanding the CC, we aim to
develop methods to mitigate it. We turn to calibration,
which has recently been used in probing with few-shot
examples by \citet{pmlr-v139-zhao21c}. The aim of
calibration is to improve the performance of a model on a
classification task, by first assessing the prior
probability of a label (i.e., its probability if no context
is given), and
then dividing the probability predicted in the task context by this
prior; this gives us  the conditional
probability of a label given the context, representing the
true knowledge of the model about this task. In adapting
calibration, we want to give a model every possible
opportunity to do well so that we do not underestimate its
underlying comprehension.

We therefore develop three different methods of removing the
important information from the context in such a way that we
can use the prediction probabilities of the two adjectives
in these contexts for calibration. The simplest way of doing
this is to remove both instances of the CC, resulting in S5
in Table \ref{semantics_schemata}. If we want to keep the CC
in the context, the two options to remove any information
are to replace either the names or the adjectives
with new names/adjectives. We therefore construct two more instances for calibration:
S6 and S7 in Table \ref{semantics_schemata}.

For each calibration method, we collect five examples with different adjectives or names. For a given base sample $S_b$, we calculate $P_c$, the calibrated predictions, as follows:
\begin{align*}
	P_c(a|S_b) = P(a|S_b)/[\sum\limits_{i=1}^{i=5}(P(a|C_i)/5)]
\end{align*}
where $C_i$ is the $i$-th example of a given calibration technique, $a$ is the list of adjectives tested for the masked position, and the division is applied elementwise.
We collect a list of 20 adjectives and their antonyms
manually from the vocabulary of the RoBERTa tokenizer and 33 common names and generate 144,800 sentences from them.
We test BERT \cite{devlin-etal-2019-BERT} in the sizes base and large, RoBERTa \cite{liu2019} in the sizes base and large, and DeBERTa \cite{he2020} in the sizes base, large, xlarge and xxlarge.

\subsection{Results}

\begin{table}[]
    \small
    \centering
        \begin{tabular}{lrrrrrr}
        \toprule
        & \multicolumn{2}{c}{Accuracy} & \multicolumn{3}{c}{Decision Flip}  \\ 
         \cmidrule(lr){2-3}  \cmidrule(lr){4-6}
         &  \multicolumn{1}{l}{S1} & \multicolumn{1}{l}{S2} & \multicolumn{1}{l}{S2} & \multicolumn{1}{l}{S3} & \multicolumn{1}{l}{S4} \\ \midrule
        BERT\textsubscript{B}     &   37.65                              & 64.64                                   & 26.98                                     & 75.69                                       & 02.70                                  \\
        BERT\textsubscript{L}    &  36.85                              & 67.21                                   & 30.44                                     & 73.31                                       & 02.32                                  \\
        RoBERTa\textsubscript{B}  &  61.60                              & 52.84                                   & 09.91                                     & 76.18                                       & 02.76                                  \\
        RoBERTa\textsubscript{L} &  55.71                              & 68.00                                   & 14.33                                     & 79.47                                       & 04.33                                  \\
        DeBERTa\textsubscript{B}  &  49.72                              & 49.80                                   & 00.91                                     & 99.66                                       & 01.07                                  \\
        DeBERTa\textsubscript{L} &  50.88                              & 51.40                                   & 07.04                                     & 94.83                                       & 02.23                                  \\
        DeBERTa\textsubscript{XL} & 47.73                              & 49.33                                   & 05.46                                     & 89.28                                       & 02.51                                  \\
        DeBERTa\textsubscript{XXL} &  47.34                              & 48.72                                   & 03.59                                     & 82.09                                       & 01.13                                  \\ \bottomrule
        \end{tabular}
    \caption{Selected accuracies and results for the
    semantic probe. We report the average accuracy on the
    more difficult sentences in terms of recency bias (S1)
    and the easier ones (S2), as well as the percentage of
    decisions flipped by changing from the base S1 to the
    sentences testing for recency bias (S2), vocabulary bias
    (S3), and name bias (S4).
RoBERTa and DeBERTa perform close to chance on S1 and S2
    accuracy, indicating that they do not understand the
    meaning of CC. BERT's performance is strongly influenced
    by biases (recency, lexical identity), also indicating
    that it has very limited if any understanding of CC.}
    \label{table_semantics}
\end{table}

In Table \ref{table_semantics}, we report the accuracy for
all examined models. Out of the three variations to test
biases, we report accuracy only for the sentence testing the
recency bias as we expect this bias to occur systematically
across all sentences: if it is a large effect, it will always
lead to the sentence where the correct answer is the more
recent one being favoured. To assess the influence of each
bias beyond accuracy, we report as decision flip the
percentage of sentences for which the decision (i.e., if the
correct adjective had a higher probability than the
incorrect one) was changed when considering the alternative
sentence that was constructed to test for bias. We report
full results in Appendix,
Table \ref{semantics_table_appendix}.

Looking at the accuracies, we see that RoBERTa's and DeBERTa's
scores are close to 50\% (i.e., chance) accuracy for
both S1 and S2. BERT models differ considerably as they seem to suffer from bias related to the order of the two CCs, but we can see that the average between them is also very close to chance. 
When we further look at the decision flips for each of the
biases, we find that there is next to no bias related to the
choice of names (S4). However, we can see a large bias
related to both the recency of the correct answer (S2) and
the choice of adjectives (S3).
The recency bias is strongest in the BERT models, which also
accounts for the difference in accuracies. For RoBERTa and
DeBERTa models, the recency bias is small, but clearly present. In contrast, they exhibit far greater bias towards the choice of adjective, even going as far as 99.66\% of decisions flipped by changing the adjective for DeBERTa base. This suggests that these models' decisions about which adjective to assign a higher probability is almost completely influenced by the choice of adjective, not the presence of the CC. 
Overall, we conclude that without calibration, all models seem to be highly susceptible to different combinations of bias, which completely obfuscate any underlying knowledge of the CC, leading to an accuracy at chance level across the board.

We therefore turn to our calibration methods, evaluating
them first on their influence on the decision flip scores,
which directly show if we were able to reduce the impact of
the different types of bias. We report these only for order
and vocabulary bias as we found name bias to be
inconsequential. We report the complete results in 
Appendix, Tables \ref{semantics_table_appendix} and \ref{decisionflip_table_appendix}. We see that across all models, while all three
calibration methods work to reduce some bias, none does so
consistently across all models or types of
bias. We report the impact of all calibration
methods on the final accuracies of the three largest models in Table \ref{calibration}. Even in
cases where calibration has clearly reduced the decision flip score, we find that the final calibrated accuracy is
still close to 50\%. This indicates that despite the
effort to retrieve any knowledge that the models
have about the CC, they are unable to perform clearly above
chance, and we have therefore found no evidence that the
investigated models
understand and can use
the semantics of the CC.

\subsubsection{Problem Analysis}
Different conclusions might be drawn as to why none of these models have learned the semantics of
the CC. They might not have seen enough examples of it
in their training corpus
to
have formed a general understanding. Given the many
examples that we were able to find in C4, and the overall
positive results from the syntax section, we find this to be
unlikely. Alternatively, it could be argued that models have
never had a chance to learn what the CC means because they
have never seen it applied, and do not have
the same opportunities as humans to either interact with
the speaker to clarify the meaning or to make deductions
using observations in the real world.
This is in line with other considerations about large PLMs acquiring advanced semantics, even though it has for many phenomena been shown that pretraining is sufficient \citep{Radford.2019}. Lastly, it might be possible that the type of meaning representation required to solve this task is beyond the current transformer-style architectures.
Overall, our finding that PLMs do not learn the semantics of the CC adds to the growing body of evidence that complex semantics like negation \cite{kassner-schutze-2020-negated} is still beyond state-of-the-art PLMs.

\section{Related Work}

\begin{table}[]
	\centering
	\small
	\begin{tabular}{llllll}
		\toprule
		Model & Test &      - &     S5 &     S6 &     S7 \\
		\midrule
		\multirow{3}{*}{BERT\textsubscript{L}} &     S1 & 36.85 & 31.91 & 47.21 & 44.03 \\
		&     S2 & 67.13 & 73.48 & 54.39 & 64.45 \\
		&     S3 & 36.46 & 43.43 & 47.79 & 44.36 \\ \midrule
		\multirow{3}{*}{RoBERTa\textsubscript{L}} &     S1 & 55.72 & 58.37 & 65.08 & 69.53 \\
		&     S2 & 68.01 & 74.53 & 62.73 & 77.76 \\
		&     S3 & 55.36 & 52.02 & 65.28 & 69.23 \\ \midrule
		\multirow{3}{*}{DeBERTa\textsubscript{XXL}}  &     S1 & 47.35 & 53.56 & 54.92 & 54.12 \\
		&     S2 & 48.73 & 52.85 & 54.03 & 53.81 \\
		&     S3 & 47.57 & 49.36 & 55.25 & 53.59 \\
		\bottomrule
	\end{tabular}
	
	\caption{Effect of our three calibration methods
        compared to no calibration, for the three largest
        models. We report the accuracy scores for the base
        sentence (S1), recency bias (S2), and vocabulary
        bias (S3).
The results indicate that, even if we try to address bias
        through calibration,
the models are unable
to perform clearly above chance. We have therefore found
no evidence that the  models understand 
the semantics of the CC.
}
	\label{calibration}
\end{table}

\subsection{Construction Grammar in NLP}
CxG has only recently and very sparsely been investigated in neural network-based NLP. 
\citet{tayyar-madabushi-etal-2020-cxgBERT} use 
a probe to show that while a probe on top of BERT contextual embeddings is able to mostly correctly 
classify if two sentences contain instances of the same construction, injecting this knowledge into the 
model by adding it to pretraining does not improve its performance. 
Our work differs from this study in that we delve deeper into what it means to understand a 
construction on a semantic level, and take careful precautions to isolate the recognition of the 
construction at the syntax level from confounding factors.
\citet{li-etal-2022-neural} recreate the experiments of \citet{bencini2000} and \citet{johnson2013} on
argument structure constructions, by creating artificial sentences with four major argument structure types
and a random combination of verbs, to investigate whether PLMs prefer sorting by construction or by main
verb. \citet{tseng-etal-2022-cxlm} choose items from a Chinese construction list and investigate PLM's
predictions when masking the open slots, the closed slots, or the entire construction. They find that
models find closed slots easier to predict than open ones. 
Other computational studies about CxG have either focused on automatically annotating constructions 
\cite{dunietz-etal-2017-automatically} or on the creation and evaluation of automatically built 
lists of constructions \citep{marques-beuls-2016-evaluation, dunn-2019-frequency}.
\subsection{Probing}

Our work also bears some similarity to recent work in
generative grammar-based syntax probing of large PLMs in
that we approximate the minimal pairs-based probing
framework
similar
to \citet{wei-etal-2021-frequency}, \citet{marvin-linzen-2018-targeted}
or \citet{goldberg2019}. However, as we are concerned with
different phenomena and investigating them from a different
theoretical standpoint, the syntactic half of our work
clearly differs.

The semantic half of our study is closest to recent work on designing challenging test cases for models such as \citet{ribeiro-etal-2020-beyond}, who design some edge cases for which most PLMs fail. Despite the different motivation, the outcome is very similar to a list of some particularly challenging constructions. 

\section{Conclusion}

We have made a first step towards a thorough investigation
of the compatibility of the paradigm of CxG and the
syntactic and semantic capabilities exhibited by
state-of-the-art large PLMs. For this, we chose the English
comparative correlative, one of the most well-studied
constructions, and investigated if large PLMs have learned
it, both syntactically and semantically. We found that even
though they are able to classify sentences as instances of
the construction even in difficult circumstances, they do
not seem to be
able to
extract the meaning it conveys 
and use it in context, indicating that while the syntactic
aspect of the CC is captured in pretraining, the semantic
aspect is not. We see this an indication that major future
work will be needed to enable neural models to fully understand language to the same degree as humans.

\section*{Limitations}
As our experimental setup requires significant customisation with regards to the 
properties of the specific construction we investigate, we are unable to consider 
other constructions or other languages in this work. We hope to be able to extend 
our experiments in this direction in the future. Our analysis is also limited---as all probing papers are---by the necessary indirectness of the probing tasks:
we cannot directly assess the model's internal representation of the CC, but only
construct tasks that might show it but are imperfect and potentially affected by 
external factors.

\section*{Acknowledgements}
This work was funded by the European Research
Council (\#740516). The second author was also supported by the German
Academic Scholarship Foundation.
The third author was also supported by the German Federal Ministry of Education and Research (BMBF, Grant No. 01IS18036A).
We thank the reviewers for their extremely helpful comments. We are also very grateful to David Mortensen and Lori Levin for 
helpful discussions and comments.

\bibliography{anthology,custom,valentin_references}
\bibliographystyle{acl_natbib}

\clearpage

\appendix

\begin{algorithm*}
\caption{Context-Free Grammar for Artificial Data Creation Training Set}\label{alg:artificial_train}
\label{cfg1}
\begin{algorithmic}
\State S $\rightarrow$ SPOS | SNEG
\State SPOS $\rightarrow$ POS1 PUNCT POS2 '.' | POS1 INSERT PUNCT POS2 '.'
\State SNEG $\rightarrow$ NEG1 PUNCT NEG2 '.' | NEG1 INSERT PUNCT NEG2 '.'
\State PUNCT $\rightarrow$ ',' | ';' | ''
\State CORE\_POS $\rightarrow$ ADV\_I 'the' NUM NOUN VERB
\State CORE\_NEG $\rightarrow$ ADV\_I NUM VERB 'the' NOUN
\State POS\_UPPER $\rightarrow$ '0 The' CORE\_POS
\State POS\_LOWER $\rightarrow$ '0 the' CORE\_POS
\State NEG\_UPPER $\rightarrow$ '0 The' CORE\_NEG
\State NEG\_LOWER $\rightarrow$ '0 the' CORE\_NEG
\State POS1 $\rightarrow$ POS\_UPPER | POS\_UPPER ADD | START POS\_LOWER | START POS\_LOWER ADD
\State POS2 $\rightarrow$ POS\_LOWER | POS\_LOWER ADD
\State NEG1 $\rightarrow$ NEG\_UPPER | NEG\_UPPER ADD | START NEG\_LOWER | START NEG\_LOWER ADD
\State NEG2 $\rightarrow$ NEG\_LOWER | NEG\_LOWER ADD
\State INSERT $\rightarrow$ INSERT1 | INSERT2
\State INSERT2 $\rightarrow$ ADDITION BETWEEN\_ADD\_AND\_SENT SENT 
\State PRON $\rightarrow$ 'we' | 'they'
\State ADDITION $\rightarrow$ ', and by the way ,' | ', and I want to add that' | ', and' PRON 'just want to say that' | ', and then' PRON 'said that' | ', and then' PRON 'said that'
\State SAY $\rightarrow$ 'say' | 'think' | 'mean' | 'believe'
\State BETWEEN\_ADD\_AND\_SENT $\rightarrow$ PRON SAY 'that' | PRON SAY 'that' | PRON SAY 'that' | PRON SAY 'that'
\State LOC\_SENT $\rightarrow$ PRON 'said this in' LOC 'too'
\State LOC $\rightarrow$ CITY 'and' LOC | CITY
\State CITY $\rightarrow$ 'Munich' | 'Washington' | 'Cologne' | 'Prague' | 'Istanbul'
\State SENT $\rightarrow$ 'this also holds in other cases' | 'this is not always true' | 'this is always true' | 'this has only recently been the case' | 'this has not always been the case' | 'this has always been the case'
\State INSERT1 $\rightarrow$ 'without stopping' | 'without a break' | 'without a pause' | 'uninterrupted' |
\State START $\rightarrow$ 'Nowadays ,' | 'Nowadays' | 'Therefore ,' | 'Therefore' | 'We can' CANWORD 'that' | 'It is' KNOWNWORD 'that' | 'It follows that' | 'Sometimes' | 'Sometimes ,' | 'It was recently announced that' | 'People have told me that' | 'I recently read in a really interesting book that' | 'I have recently read in an established , well-known newspaper that' | 'It was reported in a special segment on TV today that'
\State CANWORD $\rightarrow$ 'say' | 'surmise' | 'accept' | 'state'
\State KNOWNWORD $\rightarrow$ 'clear' | 'known' | 'accepted' | 'obvious'
\State ADD $\rightarrow$ TEMP | UNDER1 | TEMP UNDER1 | UNDER1 TEMP
\State ADV\_I $\rightarrow$ ADV | ADV 'and' ADV
\State TEMP $\rightarrow$ TEMP1 TEMP2
\State TEMP1 $\rightarrow$ 'before' | 'after' | 'during'
\State TEMP2 $\rightarrow$ 'the morning' | 'the afternoon' | 'the night'
\State UNDER1 $\rightarrow$ 'under the' UNDER2
\State UNDER2 $\rightarrow$ 'bed' | 'roof' | 'sun'
\State VERB $\rightarrow$ 'push' | 'attack' | 'chase' | 'beat' | 'believe' | 'boil' | 'box' | 'burn' | 'call' | 'date' 
\State NOUN $\rightarrow$ 'lions' | 'pandas' | 'camels' | 'pigs' | 'horses' | 'sheep' | 'chickens' | 'foxes' | 'cows' | 'deer' 
\State ADV $\rightarrow$ 'worse' | 'earlier' | 'slower' | 'deeper' | 'bigger' | 'smaller' | 'flatter' | 'weaker' | 'stronger' | 'louder' 
\State NUM $\rightarrow$ 'twelve' | 'thirteen' | 'fourteen' | 'fifteen' | 'sixteen' | 'seventeen' | 'eighteen' | 'nineteen' | 'twenty' | 'twenty-one'

\end{algorithmic}
\end{algorithm*}

\begin{algorithm*}
\caption{Context-Free Grammar for Artificial Data Creation Test Set}\label{alg:artificial_test}
\label{cfg2}
\begin{algorithmic}
\State S $\rightarrow$ SPOS | SNEG
\State SPOS $\rightarrow$ POS1 PUNCT POS2 '.' | POS1 INSERT PUNCT POS2 '.'
\State SNEG $\rightarrow$ NEG1 PUNCT NEG2 '.' | NEG1 INSERT PUNCT NEG2 '.'
\State PUNCT $\rightarrow$ ',' | ';' | ''
\State CORE\_POS $\rightarrow$ \texttt{ADV}\_I 'the' NUM \texttt{NOUN} \texttt{VERB}
\State CORE\_NEG $\rightarrow$ \texttt{ADV}\_I NUM \texttt{VERB} 'the' \texttt{NOUN}
\State POS\_UPPER $\rightarrow$ '0 The' CORE\_POS
\State POS\_LOWER $\rightarrow$ '0 the' CORE\_POS
\State NEG\_UPPER $\rightarrow$ '0 The' CORE\_NEG
\State NEG\_LOWER $\rightarrow$ '0 the' CORE\_NEG
\State POS1 $\rightarrow$ POS\_UPPER | POS\_UPPER ADD | START POS\_LOWER | START POS\_LOWER ADD
\State POS2 $\rightarrow$ POS\_LOWER | POS\_LOWER ADD
\State NEG1 $\rightarrow$ NEG\_UPPER | NEG\_UPPER ADD | START NEG\_LOWER | START NEG\_LOWER ADD
\State NEG2 $\rightarrow$ NEG\_LOWER | NEG\_LOWER ADD
\State INSERT $\rightarrow$ INSERT1 | INSERT2
\State INSERT2 $\rightarrow$ ADDITION BETWEEN\_ADD\_AND\_SENT SENT 
\State PRON $\rightarrow$ 'I' | 'you'
\State ADDITION $\rightarrow$ ', and by the way ,' | ', and I want to add that' | ', and' PRON 'just want to say that' | ', and then' PRON 'said that' | ', and then' PRON 'said that'
\State SAY $\rightarrow$ 'say' | 'think' | 'mean' | 'believe'
\State BETWEEN\_ADD\_AND\_SENT $\rightarrow$ PRON SAY 'that' | PRON SAY 'that' | PRON SAY 'that' | PRON SAY 'that'
\State LOC\_SENT $\rightarrow$ PRON 'said this in' LOC 'too'
\State LOC $\rightarrow$ CITY 'and' LOC | CITY
\State CITY $\rightarrow$ 'London' | 'New York' | 'Berlin' | 'Madrid' | 'Paris'
\State SENT $\rightarrow$ 'this also holds in other cases' | 'this is not always true' | 'this is always true' | 'this has only recently been the case' | 'this has not always been the case' | 'this has always been the case'
\State INSERT1 $\rightarrow$ 'without stopping' | 'without a break' | 'without a pause' | 'uninterrupted' |
\State START $\rightarrow$ 'Nowadays ,' | 'Nowadays' | 'Therefore ,' | 'Therefore' | 'We can' CANWORD 'that' | 'It is' KNOWNWORD 'that' | 'It follows that' | 'Sometimes' | 'Sometimes ,' | 'It was recently announced that' | 'People have told me that' | 'I recently read in a really interesting book that' | 'I have recently read in an established , well-known newspaper that' | 'It was reported in a special segment on TV today that'
\State CANWORD $\rightarrow$ 'say' | 'surmise'
\State KNOWNWORD $\rightarrow$ 'clear' | 'known'
\State ADD $\rightarrow$ TEMP | UNDER1 | TEMP UNDER1 | UNDER1 TEMP
\State \texttt{ADV}\_I $\rightarrow$ \texttt{ADV} | \texttt{ADV} 'and' \texttt{ADV}
\State TEMP $\rightarrow$ TEMP1 TEMP2
\State TEMP1 $\rightarrow$ 'before' | 'after' | 'during'
\State TEMP2 $\rightarrow$ 'the day' | 'the night' | 'the evening'
\State UNDER1 $\rightarrow$ 'under the' UNDER2
\State UNDER2 $\rightarrow$ 'bridge' | 'stairs' | 'tree'
\State \texttt{VERB} $\rightarrow$ 'slam' | 'break' | 'bleed' | 'shake' | 'smash' | 'throw' | 'strike' | 'shoot' | 'swallow' | 'choke'
\State \texttt{NOUN} $\rightarrow$ 'cats' | 'dogs' | 'girls' | 'boys' | 'men' | 'women' | 'people' | 'humans' | 'mice' | 'alligators'
\State \texttt{ADV} $\rightarrow$ 'faster' | 'quicker' | 'harder' | 'higher' | 'later' | 'longer' | 'shorter' | 'lower' | 'wider' | 'better'
\State NUM $\rightarrow$ 'two' | 'three' | 'four' | 'five' | 'six' | 'seven' | 'eight' | 'nine' | 'ten' | 'eleven'

\end{algorithmic}
\end{algorithm*}

\begin{figure*}
    \includegraphics[width=\textwidth]{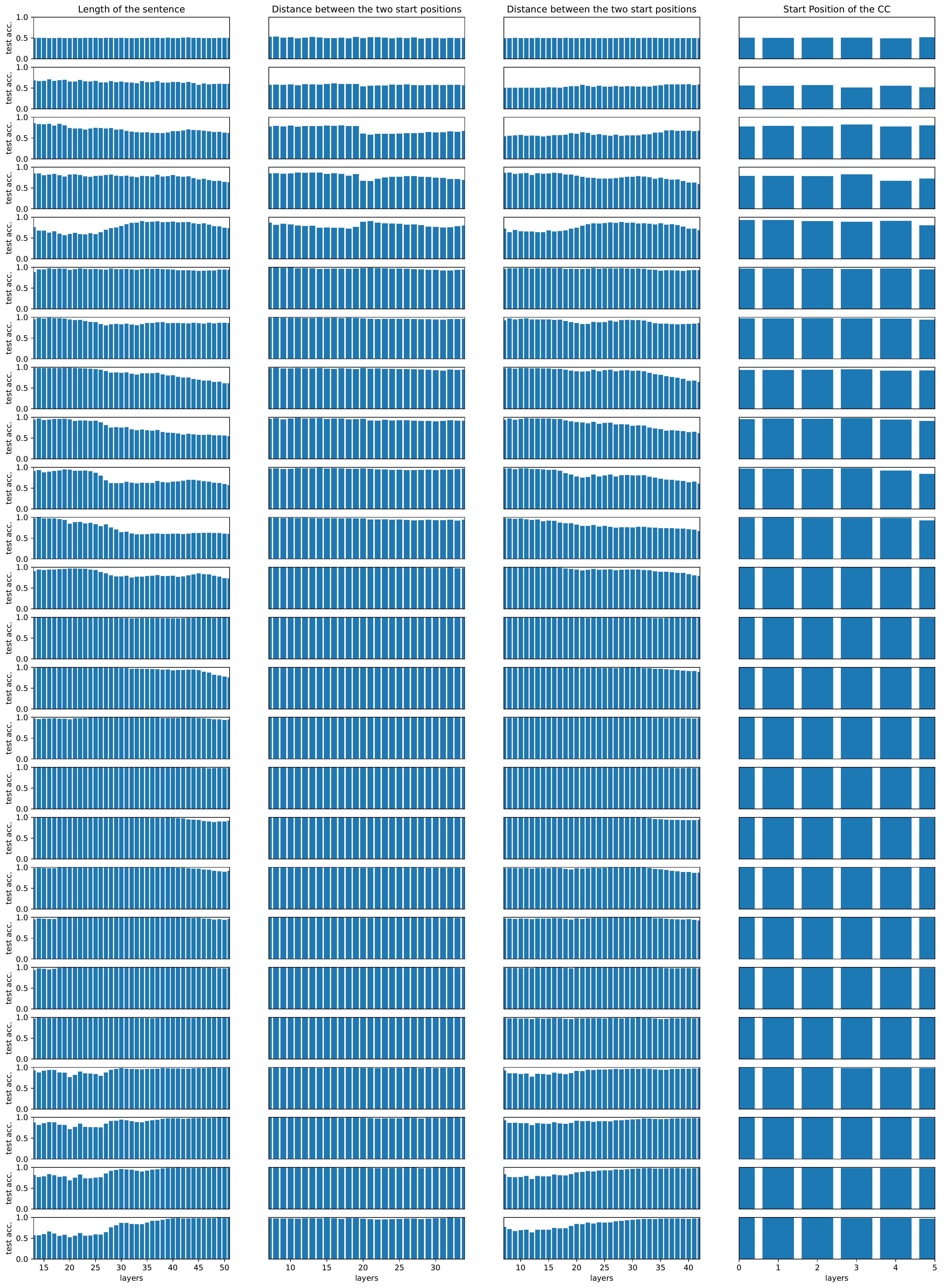}
    \caption{Full results for BERT\textsubscript{LARGE} on artificial data. Columns indicate the variable that the training and test set controls for.}
    \label{bert_artificial}
\end{figure*}

\begin{figure*}
    \includegraphics[width=\textwidth]{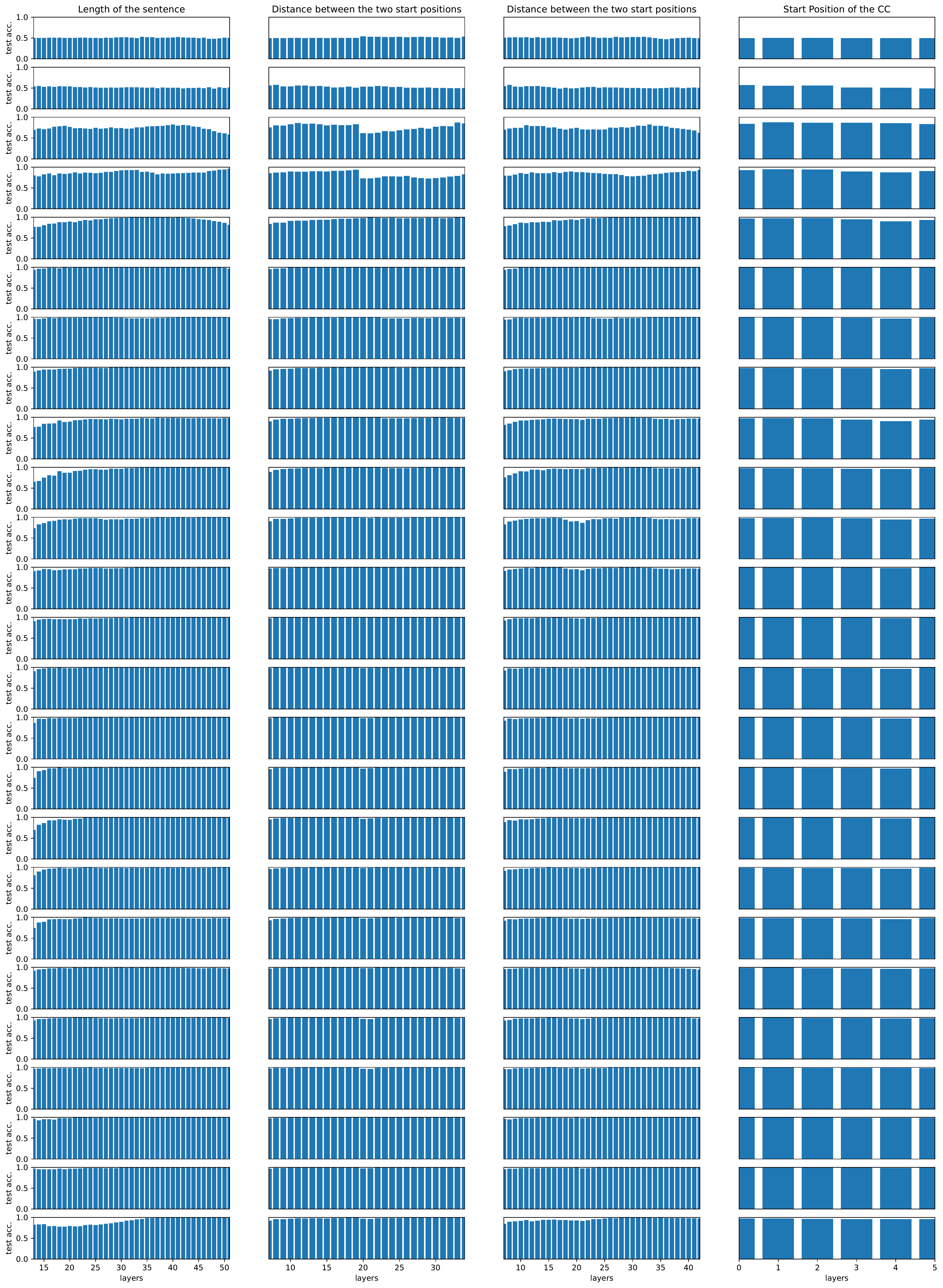}
    \caption{Full results for RoBERTa\textsubscript{LARGE} on artificial data. Columns indicate the variable that the training and test set controls for.}
    \label{roberta_artificial}
\end{figure*}

\begin{figure*}
    \includegraphics[width=\textwidth]{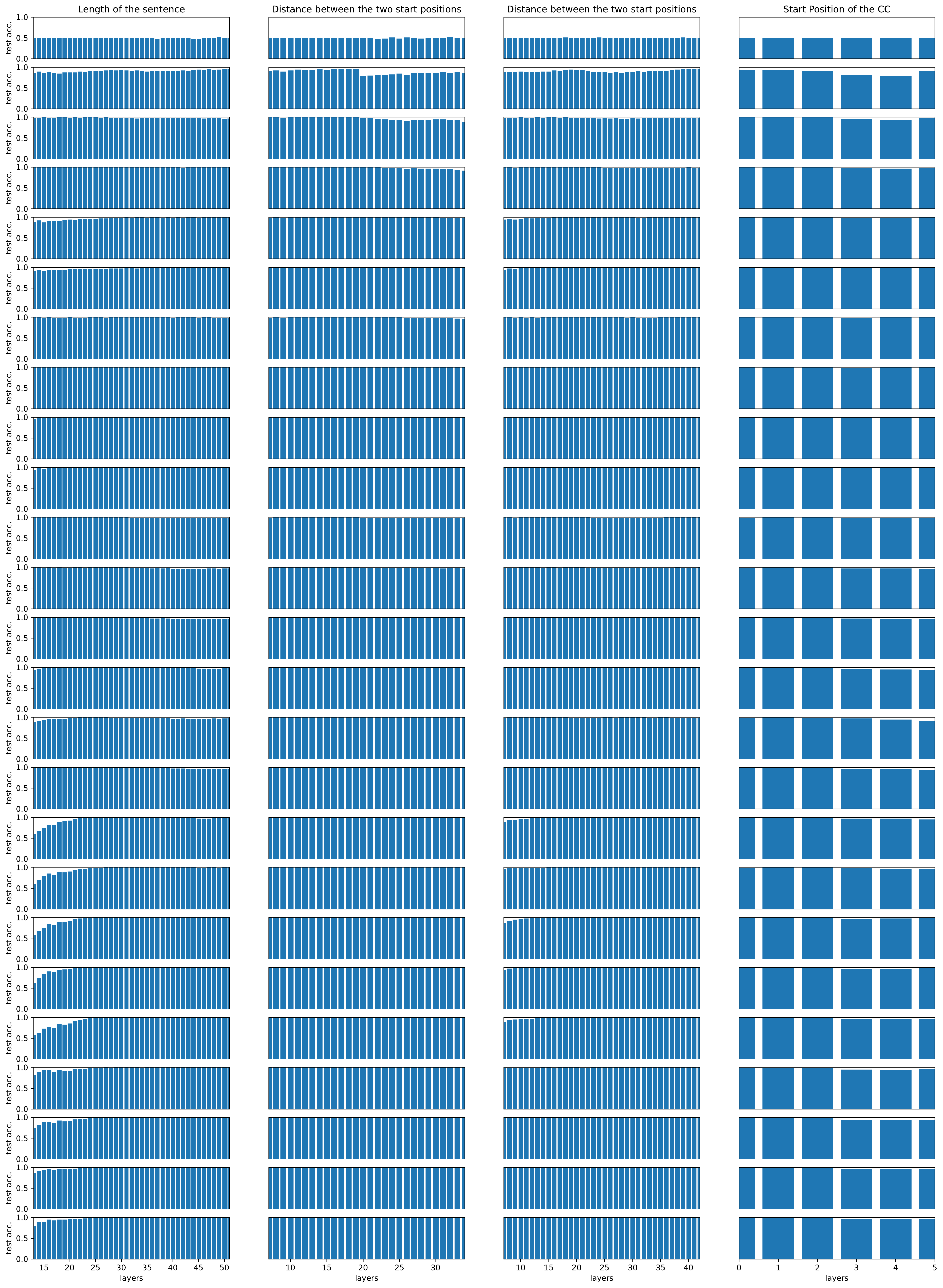}
    \caption{Full results for DeBERTa\textsubscript{LARGE} on artificial data. Columns indicate the variable that the training and test set controls for.}
    \label{deberta_artificial}
\end{figure*}

\begin{figure*}
    \includegraphics[width=\textwidth]{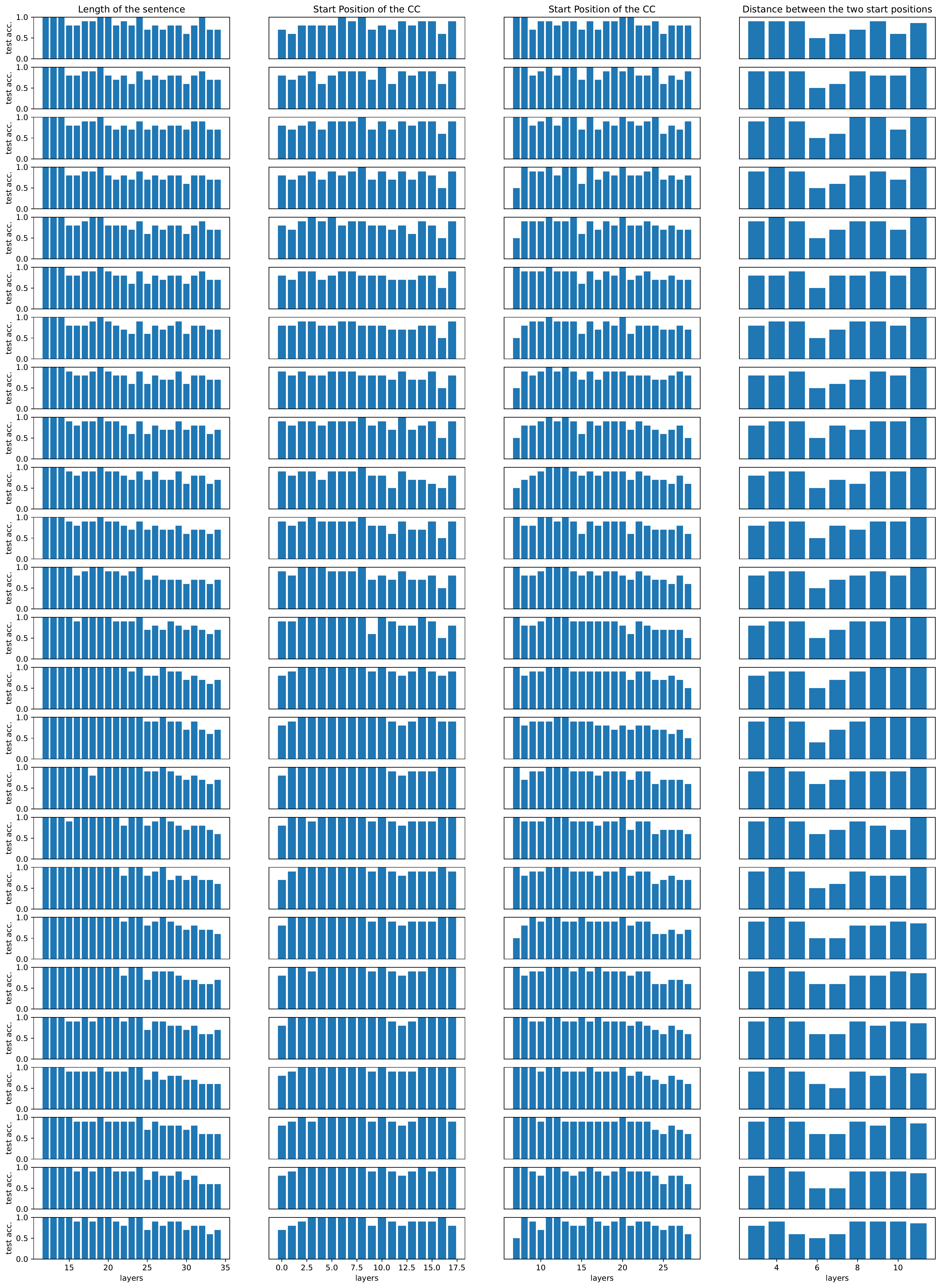}
    \caption{Full results for BERT\textsubscript{LARGE} on corpus data. Columns indicate the variable that the training and test set controls for.}
    \label{bert_corpus}
\end{figure*}

\begin{figure*}
    \includegraphics[width=\textwidth]{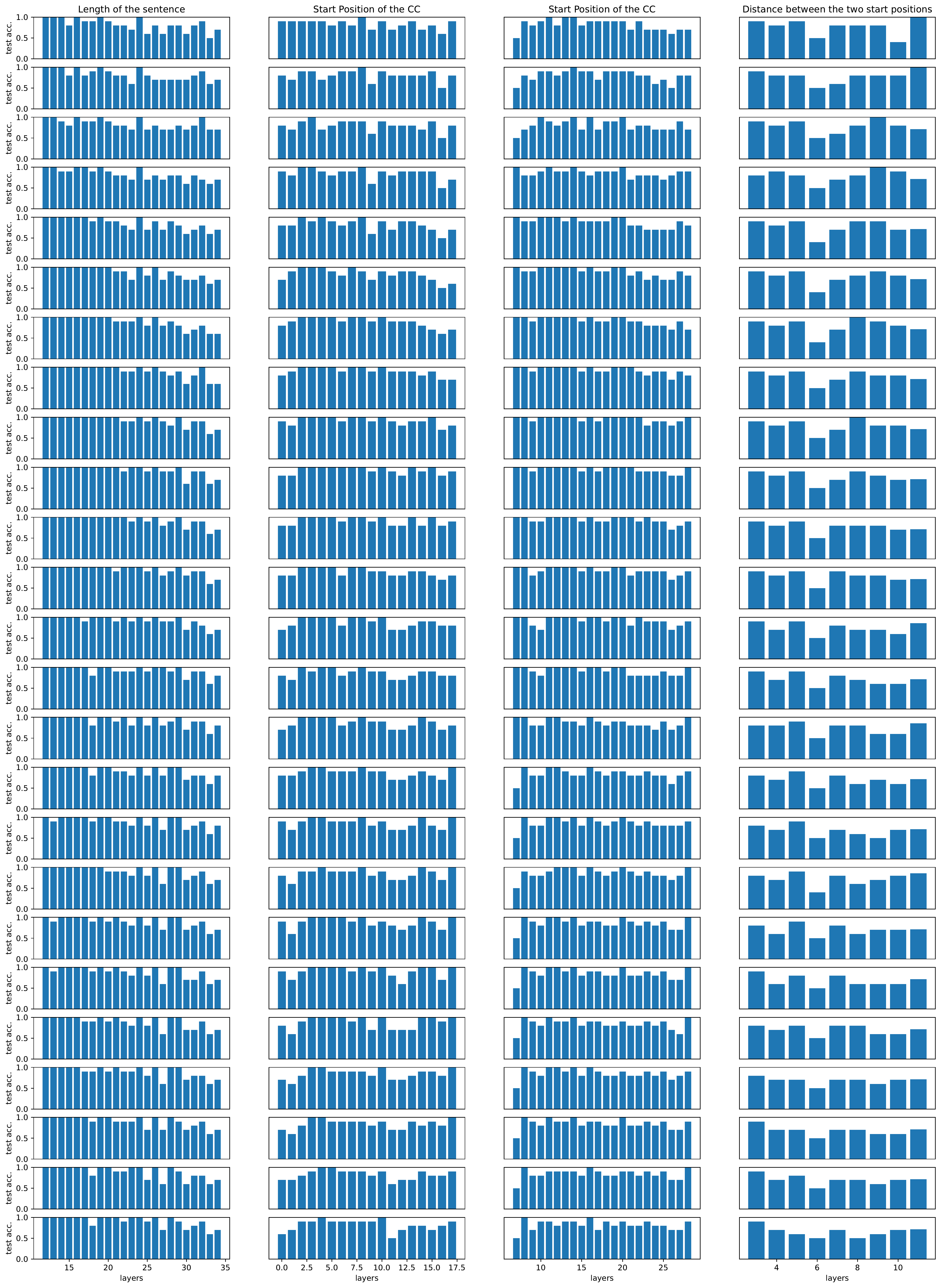}
    \caption{Full results for RoBERTa\textsubscript{LARGE} on corpus data. Columns indicate the variable that the training and test set controls for.}
    \label{roberta_corpus}
\end{figure*}

\begin{figure*}
    \includegraphics[width=\textwidth]{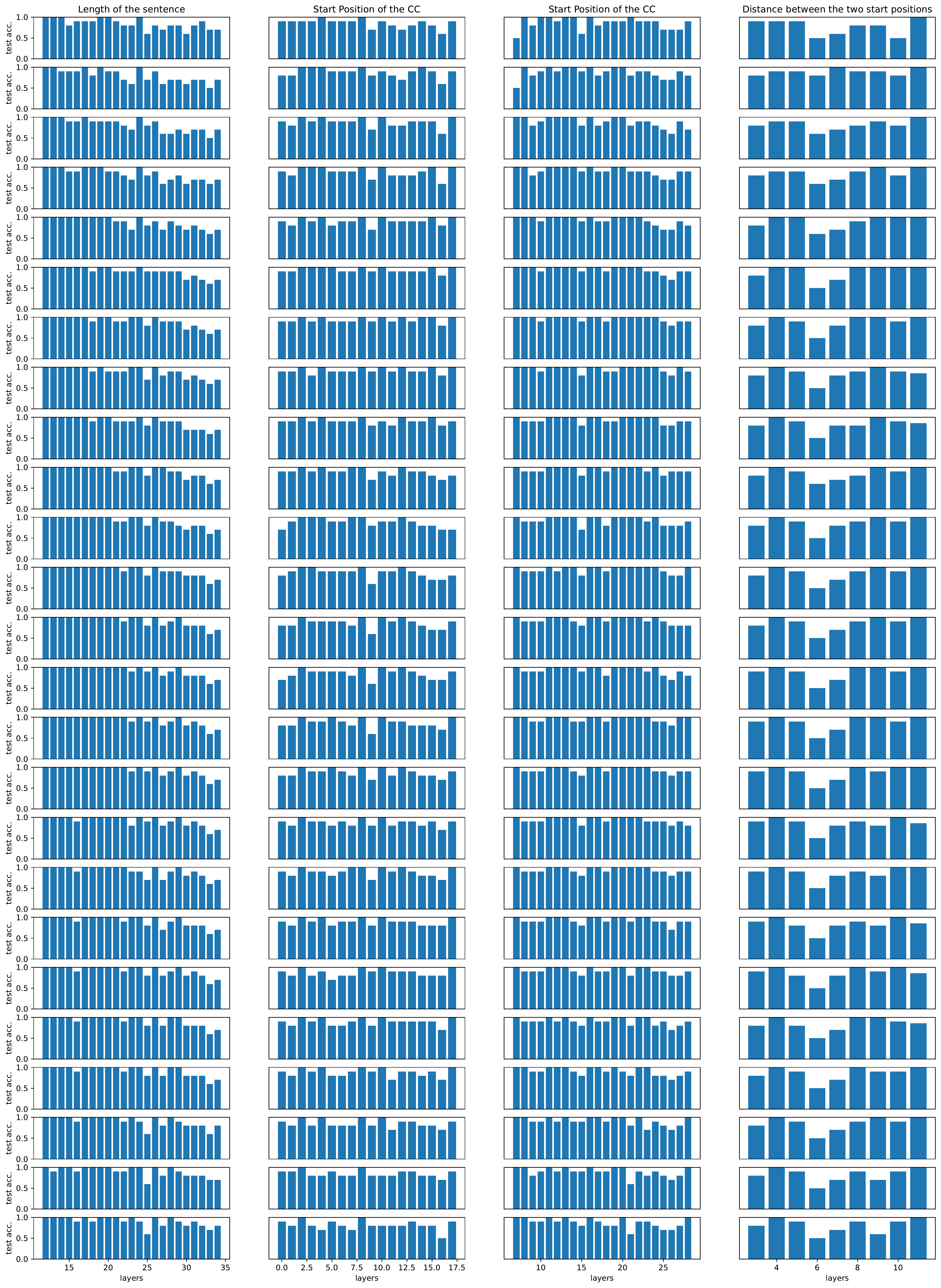}
    \caption{Full results for DeBERTa\textsubscript{LARGE} on corpus data. Columns indicate the variable that the training and test set controls for.}
    \label{deberta_corpus}
\end{figure*}

\begin{table*}
\centering
\begin{tabular}{llllll}
	\toprule
	Model & Test Scenario &      - &     S5 &     S6 &     S7 \\
	\midrule
	\multirow{3}{*}{BERT\textsubscript{B}} &     S1 & 37.65\% & 37.62\% & 44.39\% &  47.9\% \\
	 &     S2 & 64.64\% & 62.79\% & 56.66\% & 55.41\% \\
	 &     S3 & 38.04\% & 44.78\% & 44.09\% & 48.29\% \\ \midrule
	\multirow{3}{*}{BERT\textsubscript{L}} &     S1 & 36.85\% & 31.91\% & 47.21\% & 44.03\% \\
	 &     S2 & 67.13\% & 73.48\% & 54.39\% & 64.45\% \\
	 &     S3 & 36.46\% & 43.43\% & 47.79\% & 44.36\% \\ \midrule
	\multirow{3}{*}{RoBERTa\textsubscript{B}} &     S1 &  61.6\% & 58.76\% & 42.13\% & 62.32\% \\
	 &     S2 & 52.85\% & 51.35\% & 71.33\% & 60.25\% \\
	 &     S3 & 62.21\% & 55.17\% & 43.04\% & 62.76\% \\ \midrule
	\multirow{3}{*}{RoBERTa\textsubscript{L}} &     S1 & 55.72\% & 58.37\% & 65.08\% & 69.53\% \\
	 &     S2 & 68.01\% & 74.53\% & 62.73\% & 77.76\% \\
	 &     S3 & 55.36\% & 52.02\% & 65.28\% & 69.23\% \\ \midrule
	\multirow{3}{*}{DeBERTa\textsubscript{B}} &     S1 & 49.72\% & 49.72\% & 49.86\% &  49.2\% \\
	 &     S2 & 49.81\% & 48.67\% &  49.7\% & 49.06\% \\
	 &     S3 & 50.28\% & 50.19\% & 49.97\% &  50.0\% \\ \midrule
	\multirow{3}{*}{DeBERTa\textsubscript{L}}  &     S1 & 50.88\% & 49.86\% & 50.03\% & 49.39\% \\
	 &     S2 & 51.41\% & 48.09\% & 47.21\% & 48.04\% \\
	 &     S3 & 50.58\% & 49.94\% & 50.41\% & 49.42\% \\ \midrule
	\multirow{3}{*}{DeBERTa\textsubscript{XL}}  &     S1 & 47.73\% & 45.08\% & 43.31\% & 43.67\% \\
	 &     S2 & 49.34\% & 46.27\% & 45.58\% & 41.74\% \\
	 &     S3 &  47.9\% & 49.14\% & 42.68\% & 45.58\% \\ \midrule
	\multirow{3}{*}{DeBERTa\textsubscript{XXL}}  &     S1 & 47.35\% & 53.56\% & 54.92\% & 54.12\% \\
	 &     S2 & 48.73\% & 52.85\% & 54.03\% & 53.81\% \\
	 &     S3 & 47.57\% & 49.36\% & 55.25\% & 53.59\% \\
	\bottomrule
\end{tabular}

\caption{Accuracies for the semantic probe with our three calibration methods compared to no calibration. We report the average accuracy on the more difficult sentences in terms of recency bias (S1),the easier ones (S2), and vocabulary bias (S3). Our calibration tecniques are short (S5), name (S6), and adjective (S7).}
\label{semantics_table_appendix}
\end{table*}

\begin{table*}
\centering
\begin{tabular}{llllll}
		\toprule
		Model & Test Scenario &      - &     S5 &     S6 &     S7 \\
		\midrule
		\multirow{3}{*}{BERT\textsubscript{B}} &     S2 & 26.99\% & 25.22\% & 14.75\% & 10.77\% \\
		&     S3 & 75.69\% & 23.51\% & 86.33\% & 91.05\% \\
		&     S4 &  2.71\% &      - &      - &      - \\ \midrule
		\multirow{3}{*}{BERT\textsubscript{L}} &     S2 & 30.44\% &  41.8\% & 13.37\% & 22.24\% \\
		&     S3 & 73.31\% & 25.94\% & 88.65\% & 85.97\% \\
		&     S4 &  2.32\% &      - &      - &      - \\ \midrule
		\multirow{3}{*}{RoBERTa\textsubscript{B}} &     S2 &  9.92\% &  8.67\% & 31.13\% & 10.86\% \\
		&     S3 & 76.19\% & 22.04\% & 79.03\% & 74.75\% \\
		&     S4 &  2.76\% &      - &      - &      - \\ \midrule
		\multirow{3}{*}{RoBERTa\textsubscript{L}} &     S2 & 14.34\% & 17.82\% & 15.94\% & 15.86\% \\
		&     S3 & 79.48\% & 43.54\% & 64.78\% & 57.27\% \\
		&     S4 &  4.34\% &      - &      - &      - \\ \midrule
		\multirow{3}{*}{DeBERTa\textsubscript{B}} &     S2 &  0.91\% & 11.77\% &  7.13\% &  10.8\% \\
		&     S3 & 99.67\% & 56.44\% & 96.52\% & 94.94\% \\
		&     S4 &  1.08\% &      - &      - &      - \\ \midrule
		\multirow{3}{*}{DeBERTa\textsubscript{L}} &     S2 &  7.04\% &  7.85\% & 14.31\% & 14.28\% \\
		&     S3 & 94.83\% & 43.18\% & 85.75\% & 79.86\% \\
		&     S4 &  2.24\% &      - &      - &      - \\ \midrule
		\multirow{3}{*}{DeBERTa\textsubscript{XL}} &     S2 &  5.47\% &  7.87\% & 13.48\% & 18.78\% \\
		&     S3 & 89.28\% & 45.44\% & 68.48\% & 65.94\% \\
		&     S4 &  2.51\% &      - &      - &      - \\ \midrule
		\multirow{3}{*}{DeBERTa\textsubscript{XXL}} &     S2 &  3.59\% &  3.09\% & 17.02\% & 17.21\% \\
		&     S3 &  82.1\% & 79.06\% & 63.43\% & 59.81\% \\
		&     S4 &  1.13\% &      - &      - &      - \\
		\bottomrule
	\end{tabular}

\caption{Decision flip scores for the semantic probe with our three calibration methods compared to no calibration. We report the percentage of decisions flipped by changing from the base S1 to the sentences testing for recency bias (S2), vocabulary bias (S3), and name bias (S4). Our calibration tecniques are short (S5), name (S6), and adjective (S7).}
\label{decisionflip_table_appendix}
\end{table*}

\begin{table*}
\centering
    \begin{tabular}{p{150mm}|c}
    \toprule 
    \textbf{Sentence} & \textbf{Label} \\
    \midrule
         Nowadays , the bigger the eighteen sheep date , the louder and bigger the twelve horses beat under the sun . & Positive\\
         The flatter the fourteen lions push , the deeper and smaller the sixteen deer burn under the roof . & Positive\\
         The deeper the sixteen cows beat ; the flatter and earlier the twenty cows attack . & Positive\\
         Therefore , the worse the sixteen sheep believe after the morning without a pause , the smaller the thirteen cows box after the morning under the sun . & Positive\\
         The flatter the fourteen lions push , the deeper and smaller the sixteen deer burn under the roof . & Positive\\
         Sometimes , the worse and earlier seventeen believe the deer , and we just want to say that they mean that this has always been the case , the flatter twenty-one attack the foxes before the afternoon under the roof . & Negative \\
         Nowadays , the smaller sixteen box the camels , and by the way , they mean that this is always true ; the weaker thirteen date the cows . & Negative\\
         Therefore the earlier and weaker fourteen chase the deer , the stronger and earlier thirteen boil the chickens during the night . & Negative\\
         The weaker and worse fifteen box the lions during the morning under the sun , the worse twenty push the cows . & Negative\\
         It follows that the worse twelve date the pigs without a break the flatter and louder nineteen call the pigs under the sun . & Negative\\
    \bottomrule
    \end{tabular}
    \caption{Examples of artificial training data}
    \label{tab:examples_art_train}
\end{table*}

\begin{table*}
\centering
    \begin{tabular}{p{150mm}|c}
    \toprule 
    \textbf{Sentence} & \textbf{Label} \\
    \midrule
         The harder and longer the three cats throw , the harder and shorter the ten dogs shake . & Positive\\
         I have recently read in an established , well-known newspaper that the later the ten mice strike ; the later and better the seven men smash under the tree during the night . & Positive\\
         The shorter the ten girls break without a pause ; the later the ten boys bleed under the tree . & Positive\\
         It was recently announced that the better and later the five women break ; the quicker the six mice smash under the tree during the evening . & Positive\\
         The faster the seven humans choke under the stairs after the evening , and I just want to say that I think that this is not always true , the lower and higher the two boys swallow . & Positive\\
         The higher nine strike the women without a pause the shorter ten choke the girls . & Negative \\
         We can say that the longer and faster four strike the men under the stairs before the evening , the harder four throw the dogs after the day under the bridge . & Negative\\
         The quicker and higher eight bleed the people , and then I said that you believe that this also holds in other cases ; the longer seven break the girls after the night . & Negative\\
         The shorter four smash the people before the night , and by the way , you think that this is always true ; the harder three bleed the people . & Negative\\
         The longer seven shoot the women without stopping , the faster ten strike the mice after the night under the bridge . & Negative\\
    \bottomrule
    \end{tabular}
    \caption{Examples of artificial test data}
    \label{tab:examples_art_test}
\end{table*}

\begin{table*}
\centering
    \begin{tabular}{p{150mm}|c}
    \toprule 
    \textbf{Sentence} & \textbf{Label} \\
    \midrule
         " The higher up the nicer ! " & Positive\\
         She thinks the more water she drinks the better her skin looks . & Positive\\
         It becomes an obsession lightly because the more fish you catch the higher your adrenaline flows . & Positive\\
         It is worth noting , however , that the more specific you are the better . & Positive\\
         In other words , the more videos you make the greater your audience reach . & Positive\\
         Subtract the smaller from the larger . " & Negative \\
         The way the older guys help out the younger guys is fantastic . & Negative\\
         In this procedure the lower lip is pulled ventrally to expose the lower incisors . & Negative\\
         The 5th bedroom is on the lower floor with easy access to the lower bath . & Negative\\
         Note the distinctive bend of the larger vein adjacent to the smaller vein at the top . & Negative\\
    \bottomrule
    \end{tabular}
    \caption{Examples of corpus data}
    \label{tab:examples_corpus}
\end{table*}

\end{document}